\RequirePackage{fix-cm}
\documentclass[twocolumn]{svjour3}          
\smartqed  
\usepackage{graphicx}
\usepackage{algorithm}
\usepackage{algpseudocode}
\usepackage{amsmath}
\usepackage[nospace]{cite}
\usepackage{amsfonts}
\usepackage{breqn}
\usepackage{float}
\usepackage{apacite}
\usepackage[round]{natbib}
\usepackage{hyperref}
\hypersetup{
    linkcolor=blue,
    filecolor=black,      
    urlcolor=blue,
}
\algrenewcommand\algorithmicforall{\textbf{foreach}}
\algrenewcommand\algorithmicindent{.8em}

\DeclareMathOperator*{\argmax}{arg\,max}  
\DeclareMathOperator*{\argmin}{arg\,min} 
\begin{document}

\title{Robust and Adaptive Door Operation with a Mobile Robot
}


\author{Miguel Arduengo$^{1,2}$        \and
        Carme Torras$^2$            \and
        Luis Sentis$^{1,3}$  
}

\authorrunning{}

\institute{Miguel Arduengo \at
              \email{marduengo@iri.upc.edu} 
              \and
           Carme Torras \at
              \email{torras@iri.upc.edu} 
            \and
            Luis Sentis \at
            \email{lsentis@austin.utexas.edu} 
            \and
            $^{1}$Human Centered Robotics Lab (UT at Austin)\\
           $^{2}$Institut de Rob\`otica i Inform\`atica Industrial (CSIC-UPC)\\
            $^{3}$Department of Aerospace Engineering (UT at Austin)
}

\date{}

\maketitle

\begin{abstract}
The ability to deal with articulated objects is very important for robots assisting humans. In this work, a framework to robustly and adaptively operate common doors, using an autonomous mobile manipulator, is proposed. To push forward the state-of-the-art in robustness and speed performance, we devise a novel algorithm that fuses a convolutional neural network with efficient point cloud processing. This advancement enables real-time grasping pose estimation for multiple handles from RGB-D images, providing a speed up improvement for assistive human-centered applications. In addition, we propose a versatile Bayesian framework that endows the robot with the ability to infer the door kinematic model from observations of its motion and learn from previous experiences or human demonstrations. Combining these algorithms with a Task Space Region motion planner, we achieve an efficient door operation regardless of the kinematic model. We validate our framework with real-world experiments using the Toyota Human Support Robot.
\end{abstract}

\begin{figure}[t!]
\centering
\includegraphics[width=0.8\linewidth]{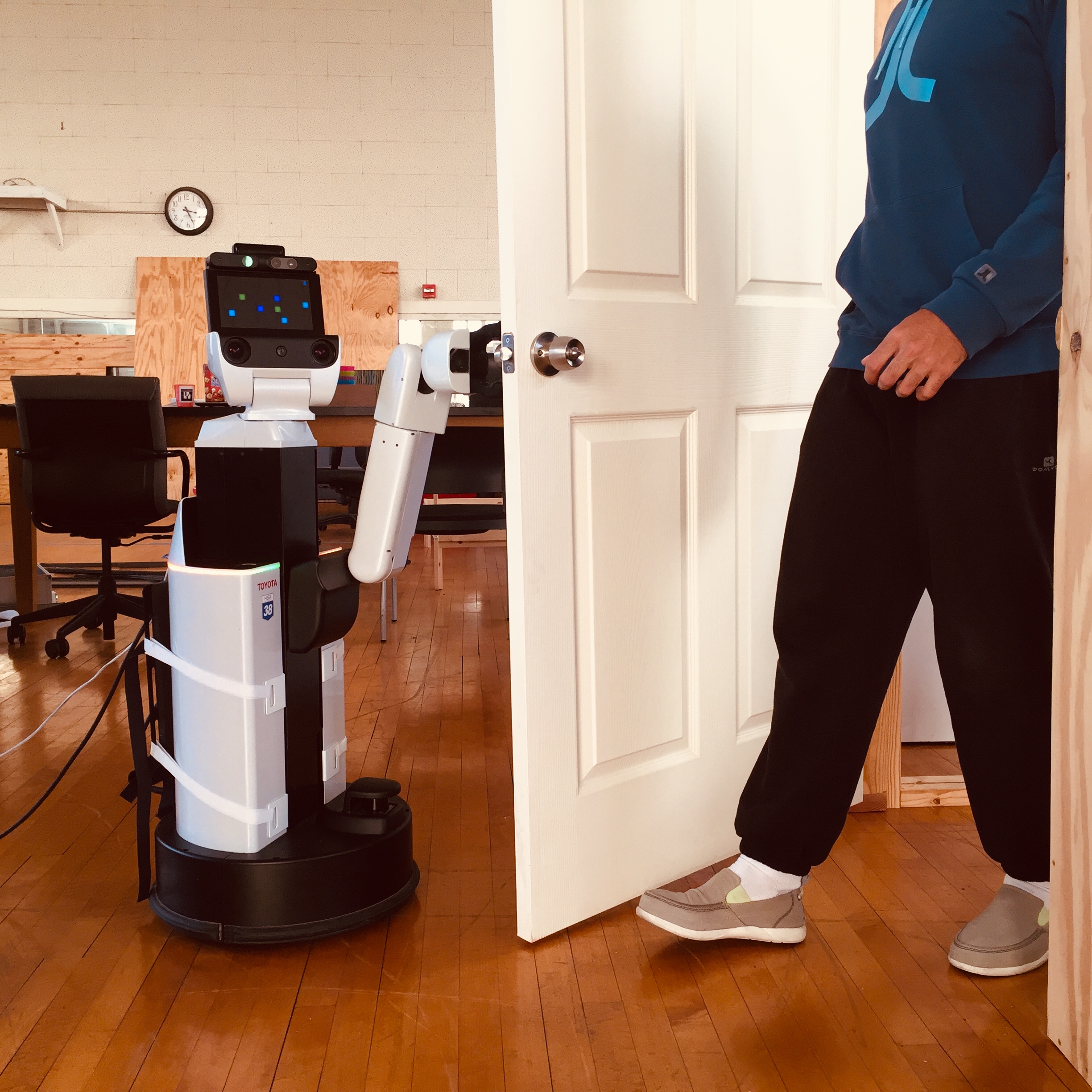}
\caption{The HSR robot assists a person to enter a room.}
\label{Fig1}
\end{figure}

\section{Introduction}
\label{intro}

\vspace{-3mm}
Robots are progressively spreading to logistic, social and assistive domains (Figure \ref{Fig1}). However, in order to become handy co-workers and helpful assistants, they must be endowed with quite different abilities than their industrial ancestors \citep{Asfour2008,Schiffer2012,Torras2016}. The ability to deal with articulated objects is relevant for robots operating in domestic environments. For example, robots need to open doors when moving around homes and to open cabinets to pick up objects \citep{Mae2011}. The problem of opening doors and drawers with robots has been tackled extensively \citep{Meeussen2010,Ott2005,Jain2009,Kessens2010,Endres2013}. These approaches usually focus either on a particular type of door and handle mechanism or on a certain aspect of the task. 

Handling different types of doors (e.g. drawers, room or refrigerator doors) and handles (e.g. doorknobs, lever handles, drawer pulls) simultaneously remains a challenge. Therefore, our contribution is on devising a more general framework that can incorporate different types of door models and that provides adaptive behavior during door operation. The paper is organized as follows: in Section \ref{related_work} we review the state-of-the-art in the field; in Section \ref{problem_statement} we state the problem addressed; in Section \ref{detection} we present our door and handle detection model; in Section \ref{grasping} we explain our approach for achieving robust real-time estimation of end-effector grasping poses; in Section \ref{unlatching} we describe a method for unlatching door handles; in Section \ref{learning} we present a Bayesian approach to learn door kinematic models which allows improving performance by learning from experience as well as from human demonstrations; in Section \ref{door_opening} we discuss the integration of kinematic model inference with a motion planner; in Section \ref{experiments} we experimentally validate our framework; finally, in Section \ref{conclusions} we draw the main conclusions.  

\section{Related Work}
\label{related_work}

The detection of doors and handles is a key problem when operating doors with an autonomous robot. A robust algorithm that allows the simultaneous detection of several doors and handles regardless of the shape, color, light conditions, etc, is essential (for instance, see Figure \ref{Fig2}). This problem has been explored based on 2D images, depth data, or both. In \citep{Chen2014}, they present a deep convolutional neural network for estimating door poses from images. Although doors are accurately located, the identification of handles is not addressed. In \citep{Banerjee2015}, following the requirements from the DARPA Robotics Challenge, the authors develop an algorithm for identifying closed doors and their handles. Doors are detected by finding consecutive pairs of vertical lines at a specific distance from one another in an image of the scene. If a flat surface is found in between, the door is recognized as closed. Handle detection is subsequently carried out by color segmentation. The paper \citep{Llopart2017} addresses the problem of detecting room doors and also cabinet doors. The authors propose a CNN to extract and identify the Region of Interest (ROI) in an RGB-D image. Then, the handle's 3D position is calculated under the assumption that it is the only object contained in the ROI and its color is significantly different from that of the door. Although positive results are obtained in these last two works, they rely on too many assumptions limiting the versatility. 

The door manipulation problem with robotic systems has also been addressed with different approaches. Some works assume substantial previous knowledge of the kinematic model of the door and its parameters, while others are entirely model-free. Among the works that assume an implicit model, in \citep{Diankov2008} the operation of articulated objects is formulated as a kinematically constrained planning problem. The authors propose to use caging grasps, to relax task constraints, and then use efficient search algorithms to produce motion plans. Another interesting work is \citep{Wieland2009}. The authors combine stereo vision and force feedback for the compliant execution of the door opening task. In recent work \cite{Eppner2018},  the authors propose candidate models that include kinematic and dynamic properties, which are selected using interactive perception. Finally, in \cite{Abraham2020} they propose a model-based path-integral controller that uses physical parameters. Regarding model-free approaches, in \citep{Lutscher2010} they propose to operate unknown doors based on an impedance control method, which adjusts the guiding speed to achieve two-dimensional planar operation. Another example is the approach presented in \citep{Karayiannidis2013}. Their method relies on force measurements and estimation of the motion direction, rotational axis and distance from the center of rotation. They propose a velocity controller that ensures a desired tangential velocity. Both approaches have their own advantages and disadvantages. By assuming an implicit kinematic model, although in practice a simpler solution is typically achieved, the applicability is limited to a single type of door. On the other hand, model-free approaches release programmers from specifying the motion parameters, but they rely entirely on the compliance of the robot. 

Alternatively, other works propose probabilistic methods that do not consider interaction forces. In \citep{Nemec2017} the authors combine reinforcement learning with intelligent control algorithms. With their method, the robot is able to learn the door-opening policy by trial and error in a simulated environment. Then, the skill is transferred to the real robot. In \citep{Welschehold2017} the authors present an approach to learn door opening action models from human demonstrations. The main limitation of these works is that they do not allow to operate autonomously unknown doors. Finally, the probabilistic approach proposed in \citep{Sturm2013} enables the description of the geometric relation between object parts to infer the kinematic structure from observations of their motion. We have adopted this approach as a basic reference but extended its capabilities to improve the performance by using prior information or human demonstrations.  

In this paper, we propose a robust and adaptive framework for manipulating general types of door mechanisms. We consider all the stages of the door opening task in a unified framework. The main contributions of our work are (a) the development of a novel algorithm to estimate the robot\textquoteright s end-effector grasping pose in real-time for multiple handles simultaneously; (b) a versatile framework that provides the robust detection and subsequent door operation for different types of door kinematic models; (c) the analysis of the door kinematic inference process by taking into account door prior information; (d) the testing on real hardware using the Toyota HSR Robot (Figure \ref{Fig1}). 

\section{Problem Statement and Framework Overview}
\label{problem_statement}

We study the problem of enabling a robot to open doors autonomously, regardless of their form or kinematic model. Performing this task requires the exploitation of the robot's sensorial, actuation and computational capabilities. The following sub-tasks are to be performed sequentially by the autonomous robot:
\begin{enumerate}
    \item \textbf{Door and handle detection} (Section \ref{detection}): First, the door and handle must be identified and located in the environment where the robot is operating. A vision system that allows the recognition of the corresponding Regions of Interest is required.
    
    \vspace{1mm}
    
    \item \textbf{Grasping of the handle} (Section \ref{grasping}): Once the handle is located, the robot must position and orient the end-effector adequately in order to grasp it. For inferring this pose, a vision system that also provides information about the 3-dimensional structure of the environment is needed.
    
    \vspace{1mm}
    
    \item \textbf{Unlatching the handle} (Section \ref{unlatching}): Then, the robot must exert an appropriate torque for actuating the handle mechanism. For determining such torque, a sensor that provides force feedback in the robot's end-effector is essential.
    
    \vspace{1mm}
    
    \item \textbf{Estimating the door kinematic model} (Section \ref{learning}): The robust operation of doors involves the inference of the required opening motion, i.e. the door kinematic model. This inference can be either performed online while actuating the door, or from observations of the door motion provided by a teacher. 
    
    \vspace{1mm}
    
    \item \textbf{Planning and executing door opening motion} (Section \ref{door_opening}): Finally, the control actions for opening the door according to its kinematic model must be planned and executed. For a mobile manipulator robot, this implies tight base-arm coordination under the task constraints.
\end{enumerate}

In this paper, we assume the robot is equipped with a vision system able to capture features in a 3-dimensional space. Additionally, we consider the particular case of a mobile robot with an omnidirectional base and that force/torque feedback in the end-effector is available. Note that these assumptions attempt to be as general as possible, as these requirements are usually met by most service robots nowadays. 

The proposed framework is structured sequentially following the sub-tasks scheme discussed above. Rather than having a distinct contribution to a specific detailed theory or methodology itself, in this work, we combine several state-of-the-art studies. Therefore, the reader can directly refer to the section of interest, indicated in the aforementioned list. The most novel approaches for addressing the door opening task are those presented for sub-tasks 2, 4 and 5.

\section{Door and Handle Detection}
\label{detection}

Doors and handles present a wide variety of geometries, sizes, colors, etc. Thus, a robust detection algorithm is essential. Additionally, in order to achieve real-time estimation, it must operate at speeds of several frames-per-second (fps). Object detection is the task of simultaneously classifying and localizing multiple objects in an image. In \citep{Redmond2016a} the authors proposed the You Only Look Once (YOLO) algorithm, an open-source state-of-the-art object detector with CNN-based regression. This network uses features from the entire image to predict each bounding box, reasoning globally about the full image and all the objects in the image. It enables end-to-end training and facilitates real-time speeds while maintaining high average precision. For these reasons, we decided to adopt this CNN architecture and train it with a custom dataset for addressing the door and handle detection problem. 

\vspace{-3mm}
\subsection{Model Training}

\begin{figure}[h]
\centering
\includegraphics[width=0.95\linewidth]{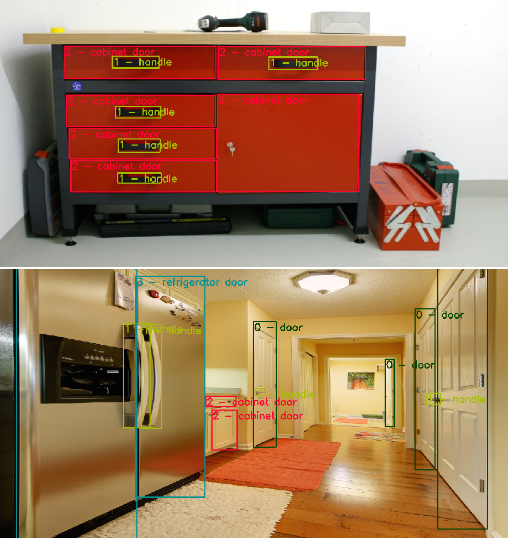}
\caption{Examples of annotated images from the training dataset used for building the door and handle detection model. The bounding boxes enclose the objects, with the corresponding label, that should be identified by the model.}
\label{Fig2}
\end{figure}

Training the YOLO network with a custom dataset allows us to build a handle and door detection model. The simplest classification semantics for our objects of interest are ``door'' and ``handle''. However, to increase the detail of the information and also to make our method versatile and extendable to other applications, we propose to split the class door into three classes: ``door'', which refers to a room door, ``cabinet door'', which includes all sorts of small doors such as drawers or a locker door, and ``refrigerator door''. We built a data set using images from the Open Images Dataset \citep{Kuznetsova2020} and annotated a total of $1213$ images containing objects of our desired object classes.

A total of $1013$ images were used for the training set, and the remaining $200$ for the testing set (the dataset is available in \citep{GitHub2019}). Some examples of the annotated images are shown in Figure \ref{Fig2}. We also applied data augmentation techniques to improve the generalization capabilities \citep{Taylor2018}. 

\vspace{-3mm}
\subsection{Model Selection}

For selecting the CNN weights and assessing the model quality, we applied cross-validation against the test set. As the performance index, we propose to use the mean average precision (mAP). This criterion was defined in the PASCAL VOC 2012 competition and is the standard metric for object detectors \citep{Everingham2015}. Briefly, the mAP computation involves the following steps: (1) Based on the likelihood of the predictions, a precision-recall curve is computed for each class, varying the likelihood threshold. (2) The area under this curve is the average precision. Averaging over the different classes we obtain the mAP. Precision and recall are calculated as:\begin{equation}
\centering
\text{Precision}=\frac{TP}{TP+FP}\quad\quad\text{Recall}=\frac{TP}{TP+FN}    
\end{equation} where $TP=$ True Positive, $TN=$ True Negative, $FP=$ False Positive and $FN=$ False Negative. True or false refers to the assigned classification being correct or incorrect, while positive or negative refers to whether the object is assigned or not to a category.

\section{Grasping the Handle}
\label{grasping}

When a robot moves towards an object, it is actually moving towards a pose at which it expects the object to be. For solving the grasping problem, the handle's 6D-pose estimation is essential. The end-effector grasping goal pose can be then easily expressed relative to the handle's pose, and reached by solving the inverse kinematics of the robot. Perception is usually provided by means of an RGB-D sensor, which supplies an RGB image and its corresponding depth map \citep{Alenya2014, Elbasiony2018}. For estimating the 6-D pose in real-time, we propose to: (1) Identify the region of the RGB image where the door and the handle are located. (2) Filter the RGB-D image to extract the Regions of Interest, clean the noise and downsample. (3) From a set of 3D geometric features of the door and the handle, estimate the grasping pose. We explain in detail these steps in this section. The proposed approach is summarized in the algorithm below:

\begin{algorithm}
\caption{\bf End-Effector Grasping Pose Estimation}
\small
\label{Algorithm1}
\begin{algorithmic}

 \renewcommand{\algorithmicrequire}{\textbf{Input:}}
 \renewcommand{\algorithmicensure}{\textbf{Output:}}

 \Require RGB image $\mathcal{I}$ and point cloud $\mathcal{P}= \left \{ \textbf{p}_j \right \} ^{N_{points}}_{0}$ 
 
 \Ensure Grasping poses $\mathcal{G}=\left \{ \textbf{g}_k \right \}^{N_{handles}}_1$ with $\textbf{g}_k \in SE(3) $

 \State Bounding boxes $\mathcal{B}=\left \{ b_l\right \}^{N_{objects}}_1 \gets \text{Detect\_Objects}(\mathcal{I})$

 \ForAll{$b_l\in \mathcal{B}$}

   \State $\mathcal{P}^{ROI}_{l} \gets \text{ROI\_Segmentation}(\mathcal{P})$

   \State $\mathcal{P}^{denoised}_{l} \gets \text{Remove\_Statistical\_Outliers}(\mathcal{P}^{ROI}_{l})$

   \State $\mathcal{P}^{filtered}_{l} \gets \text{Downsample}(\mathcal{P}^{denoised}_{l})$

   \If {$\text{class}(b_l)=\text{"handle"}$}
 
     \State $\text{orientation}_l \gets \text{Bounding\_Box\_Dimensions}(b_l)$ 
      
     \State $\mathcal{P}^{handle}_{l} \gets \text{RANSAC\_Plane\_Outliers}(\mathcal{P}^{ROI}_{l})$
      
     \State $\textbf{O}_l \gets \text{Centroid}(\mathcal{P}^{handle}_l)$
   
  \Else
  
   \State Normal $\textbf{a}_l; \, \mathcal{P}^{door}_{l} \gets \text{RANSAC\_Plane}(\mathcal{P}^{filtered}_{l})$
   
   \State $\textbf{O}_l \gets \text{Centroid}(\mathcal{P}^{door}_l)$
   
  \EndIf
 \EndFor

\State $k=1$

\ForAll{$b_l\in \mathcal{B}$ that $\text{class}(b_l)="\text{handle}"$}

  \State $\textbf{a}_l \gets \text{Assign\_Closest\_Door}(\textbf{O}_l)$

  \State $\textbf{h}_k \in SE(3) \gets \text{Handle\_Transform}(\textbf{a}_l \, ; \, \textbf{O}_l)$

  \State $\textbf{g}_k \gets \text{Goal\_Pose}(\textbf{h}_k\, ; \, \text{orientation}_l)$

  \State $k \gets k+1$\
\EndFor
\newline
\Return $\mathcal{G}$

\end{algorithmic}
\end{algorithm}

\vspace{-5mm}
\subsection{Point Cloud Filtering}

Raw point clouds contain a large number of point samples, but only a small fraction of them are of interest. Furthermore, they are unavoidably contaminated with noise. Point cloud data needs to be filtered adequately for achieving accurate feature extraction and real-time processing. We propose the following filtering process:

\subsubsection{Regions Of Interest (ROIs) Segmentation}

\begin{figure}
\centering
\includegraphics[width=1.0\linewidth]{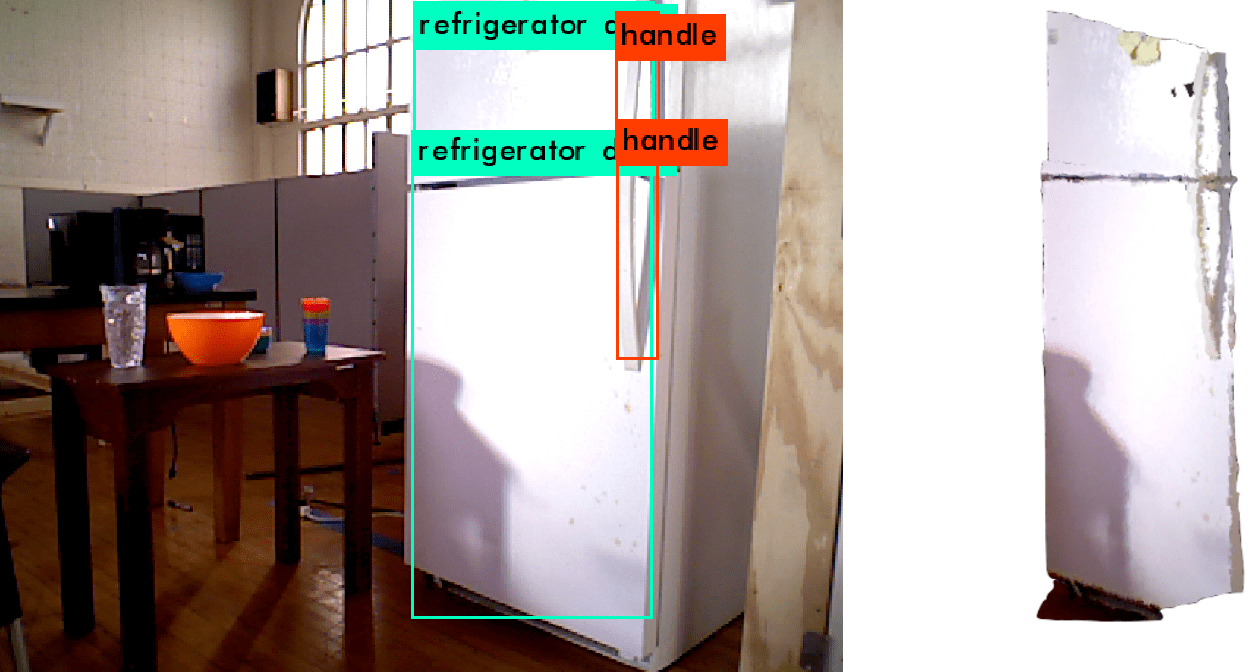}
\caption{On the left, the Regions of Interest detected by our door and handle detection model. On the right, the subset of the corresponding point cloud enclosed in the ROIs.}
\label{Fig3}
\end{figure}

The points of interest correspond to the doors and the handles in the scene, which can be defined as those in the bounding boxes of the object detection CNN. By separating the sets of points in each ROI, the amount of data to be processed is reduced significantly (Figure \ref{Fig3}). There is a direct correspondence between the pixels in the image and the point cloud indexes if the latter is indexed according to its spatial distribution. As the bounding boxes are usually provided in pixel coordinates, let $\mathcal{P}$ be the raw point cloud. Then, each ROI can be defined as follows:
\begin{equation}
\mathcal{P}^{ROI}=\left\{ \textbf{p}_{j}\in\mathcal{P}\,\vert\,j=\text{width}\cdot y+x\right\} 
\end{equation}
where $j$ is the point cloud index; $\text{width}$ is the image width in pixels, $x\in\left[x_{min},\;x_{max}\right]$ and $y\in\left[y_{min},\;y_{max}\right]$, being $\left(x_{min},y_{min}\right)$ and $\left(x_{max},y_{max}\right)$ two opposite corners of the bounding box in pixel coordinates.

\subsubsection{Statistical Outlier Filtering}

Measurement errors lead to sparse outliers, which complicate the estimation of local point cloud features such as surface normals. Some of these irregularities can be solved by performing statistical analysis of each point neighborhood, and trimming those that do not meet a certain criterion. We can carry this analysis at a discrete point level. By assuming that the average distance from every point to all its neighboring points $r_j$, can be described by a Gaussian distribution, the filtered point cloud can be defined as follows:
\begin{equation}
    \centering
    \mathcal{P}^{denoised}=\left\{ \mathbf{p}_{j}\in\mathcal{P}^{ROI}\mid r_{j}\in\left[\mu_r\pm\alpha\cdot\sigma_r\right]\right\}
\end{equation} where $\alpha$ is a multiplier, and $\mu_{r}$ and $\sigma_{r}$ are the mean distance and the standard deviation, respectively.

\subsubsection{Downsampling}

\begin{figure}
\centering
\includegraphics[width=1.0\linewidth]{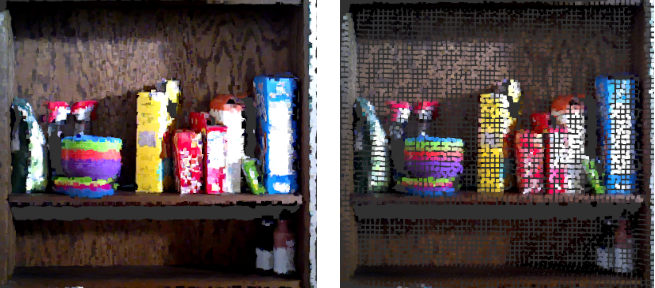}
\caption{On the left, the raw point cloud. On the right, the downsampled point cloud using a voxelized grid approach.}
\label{Fig4}
\end{figure}

In order to lighten up the computational load we propose to reduce considerably the amount of data by using a voxelized grid approach (Figure \ref{Fig4}). Unlike other sub-sampling methods, the shape characteristics are maintained. If $s$ is the number of points contained in each voxel $A$, the set of points in each voxel is replaced by:
\begin{equation}
   \bar{x}=\frac{1}{s}\sum_{A}x\quad\quad\bar{y}=\frac{1}{s}\sum_{ A}y\quad\quad\bar{z}=\frac{1}{s}\sum_{A}z 
\end{equation}

\vspace{-3mm}

\vspace{-6mm}
\subsection{Grasping Pose Estimation}

We have considered three geometric features of the 3D structure of the door and the handle for the grasping pose estimation: the handle orientation, its position and the door plane normal direction.

\subsubsection{Handle Orientation}

The end-effector orientation for grasping the handle depends on this feature. Since door handles are commonly only oriented vertically or horizontally (for a doorknob, full orientation is not relevant to grasp it), the binary decision can be made by comparing the lengths of the sides of the bounding boxes for the handles in the output of the CNN. If the height is greater than the width, the handle orientation will be vertical and vice versa.

\subsubsection{Door Plane Normal\label{subsec:Door=002019s-Plane-Normal}}

In order to grasp the handle correctly, the normal to the ``palm'' of the robot's end-effector (which we consider similar to the human hand) must be parallel to the door normal. We propose to use the RAndom SAmple Consensus (RANSAC) algorithm \citep{Rusu2013} to compute the normal direction. RANSAC is a numerical method that can iteratively estimate the parameters of a given mathematical model from experimental data that contains outliers, in such a way that they do not influence the values of the estimates. 

A minimal set is formed by the smallest number of points required to uniquely define a given type of geometric primitive. The resulting candidate shapes are tested against all points in the data to determine how many of the points are well approximated by the primitive. RANSAC estimates the model by maximizing the number of inliers \citep{Zuliani2012}. 

\begin{figure}[t]
\centering
\includegraphics[width=0.9\linewidth]{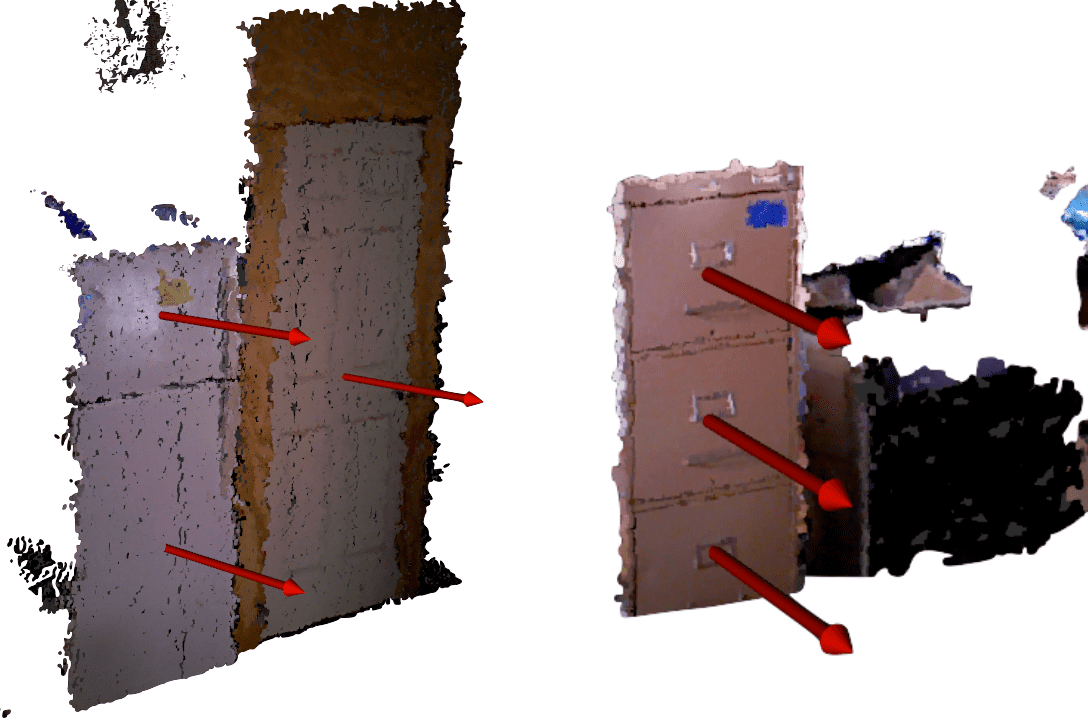}
\caption{The red arrows show the normal direction of the plane defined by each detected door.}
\label{Fig5}
\end{figure}

Then, in order to compute the door normal direction we fit a planar model to the door point cloud and calculate the coefficients of its parametric Hessian normal form using RANSAC. In Figure \ref{Fig5} we show some examples of the resulting normal vectors obtained with RANSAC. 

\subsubsection{Handle Position}

\begin{figure}[t]
\centering
\includegraphics[width=0.9\linewidth]{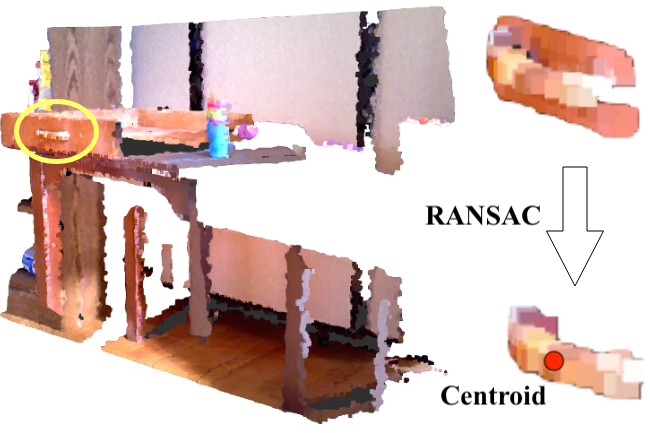}
\caption{Computation of the handle position. On the left, the observed scene with the handle highlighted. On the upper-right corner the ROI. On the lower-right corner, the ROI is filtered using RANSAC and the handle position is obtained as the centroid of the resulting point cloud.}
\label{Fig15}
\end{figure}

The proposed approach for estimating the handle position is illustrated in Figure \ref{Fig15}. We make the assumption that the handle position can be represented by its centroid. However, it cannot be directly computed from the sub-point cloud associated with the handle ROI, since the defining bounding box usually may include some points from the door in the background. Then, we also use the RANSAC algorithm to separate these points. By fitting a planar model, the ROI points can be classified as inliers and outliers. In this case, the outlier subset corresponds to the handle. The position can then be computed as the centroid of the outliers subset.

\subsubsection{Goal Pose Generation}

Let $\textbf{O}=\left(O_{x},O_{y},O_{z}\right)$ be the handle centroid and $\mathbf{a}=\left(a_{x},a_{y},a_{z}\right)$ the door plane normal unitary vector, both expressed in an arbitrary reference frame $w$. The  handle pose can be defined as the following transform:
\begin{equation}
    \small{\mathbf{T}_{w}^{handle}=\left(\begin{array}{cccc}
a_{x} & \frac{a_{y}}{a_x^2+a_y^2} & \frac{a_{x}a_{z}}{a_x^2+a_y^2} & O_{x}\\
a_{y} & -\frac{a_{x}}{a_x^2+a_y^2} & -\frac{a_{y}a_{z}}{a_x^2+a_y^2} & O_{y}\\
a_{z} & 0 & -1 & O_{z}\\
0 & 0 & 0 & 1
\end{array}\right)}
\end{equation}

\begin{figure*}
\centering
\includegraphics[width=0.98\linewidth]{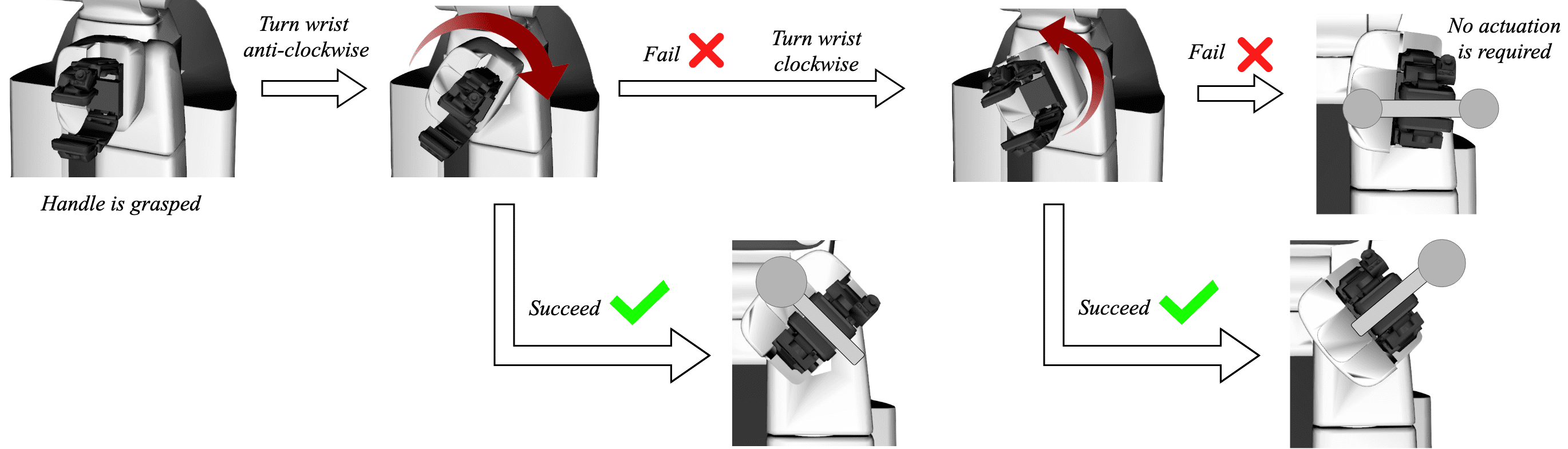}
\caption{Proposed handle unlatching strategy. First, the wrist is turned anti-clockwise. If torque feedback is above the allowed threshold, the movement is aborted. Then, the wrist is turned clockwise. If torque feedback is also above the threshold, the handle is identified as ``no actuation is required''.}
\label{Fig6}
\end{figure*}

The grasping pose can then be easily specified as a relative transform to the handle reference frame $\mathbf{T}_{\textit{handle}}^{\textit{grasping}}$, taking into account its orientation. Thus, the pose for which the Inverse Kinematics (IK) of the robot must be solved in order to finally grasp the handle, can be computed as:
\begin{equation}
    \mathbf{T}_w^{\textit{grasping}}=\mathbf{T}_w^{\textit{handle}}\mathbf{T}_{\textit{handle}}^{\textit{grasping}}
\end{equation}

\section{Unlatching the Handle}
\label{unlatching}

There exists a variety of mechanisms to open a door. Some of them do not require any specific actuation while others generally require a rotation to be applied. A handle usually occupies a small region in the door image. Thus, in order to estimate its kinematic model from visual perception data, the robot camera needs to be placed very close to the handle. This operation would increase considerably the time required to perform the task, making the inference of the handle model using the vision system unappealing. 

Instead, we rely on force feedback on the robot's end-effector for inferring how the handle should be actuated. We propose a simple, trial-and-error strategy for operating different types of handle mechanisms, illustrated in Figure \ref{Fig6}. The robot tries to turn the handle in both directions, first anti-clockwise and then clockwise. Depending on a torque threshold, either the door is unlatched or no actuation is required. Similarly, as people proceed when opening a door, using the force readings in the direction perpendicular to the door we can determine whether it is required to pull or push to open it by trial-and-error.

\vspace{-3mm}
\section{Learning the Door Kinematic Model}
\label{learning}

Opening doors in unstructured environments is challenging for robots because they have to deal with uncertainty since the kinematic model of the door is not known a priori. What if a robot has no previous knowledge of the door at the time of taking a decision? And, what if previous knowledge is available? To address these questions, we will present a probabilistic framework that allows inferring the kinematic model of the door when no previous knowledge is available and improve the performance based on previous experiences or human demonstrations.

\subsection{Overview of the Probabilistic Framework}

Let $\mathcal{D}=(\mathbf{d}_{1},\dots,\mathbf{d}_{N})$ be the sequence of $N$ relative transformations between an arbitrary fixed reference frame and the door, observed by the robot. We assume that the measurements are affected by Gaussian noise and, also, that some of these observations are outliers but not originated by the noise. Instead, the outliers might be the result of sensor failures. We denote the kinematic link model as $\mathcal{M}$. Its associated parameters are contained in the vector $\boldsymbol{\theta}\in\mathbb{R}^{k}$ (where $k$ is the number of parameters). The model that best represents the data can be formulated in a probabilistic context as \citep{Sturm2010}:
\begin{equation}
    (\hat{\mathcal{M}},\hat{\boldsymbol{\theta})}=\argmax_{\mathcal{M},\boldsymbol{\theta}}{p\left(\mathcal{\mathcal{M}},\boldsymbol{\theta}\mid\mathcal{D}\right)}
\end{equation}

This optimization is a two-step process \citep{MacKay2003}. First, a particular model is assumed true and its parameters are estimated from the observations:

\begin{equation}
    \hat{\boldsymbol{\theta}}=\argmax_{\boldsymbol{\theta}}p\left(\boldsymbol{\theta}\mid\mathcal{D},\mathcal{M}\right)
\end{equation} 

By applying Bayes rule, and assuming that the prior over the parameter space is uniform, this is equivalent to:

\begin{equation}
   \hat{\boldsymbol{\theta}}=\argmax_{\boldsymbol{\theta}}{p(\mathcal{D}\mid\boldsymbol{\theta},\mathcal{M}}) 
\end{equation} which shows that fitting a link model to the observations is equivalent to maximizing the data likelihood. Then, we can compare the probability of different models, and select the one with the highest posterior probability:

\begin{equation}
    \hat{\mathcal{M}}=\argmax_{\mathcal{M}}{\int p\left(\mathcal{M},\mathbf{\boldsymbol{\theta}}\mid\mathcal{D}\right)d\boldsymbol{\theta}}
\end{equation}

Summarizing, given a set of observations $\mathcal{D}$, and candidate models $\mathcal{M}$ with parameters $\boldsymbol{\theta}$, the procedure to infer the kinematic model of the door consists of: (1) fitting the parameters of all candidate models; (2) selecting the model that best describes the observed motion.

\subsection{Candidate Models}

\begin{figure}
\centering
\includegraphics[width=0.98\linewidth]{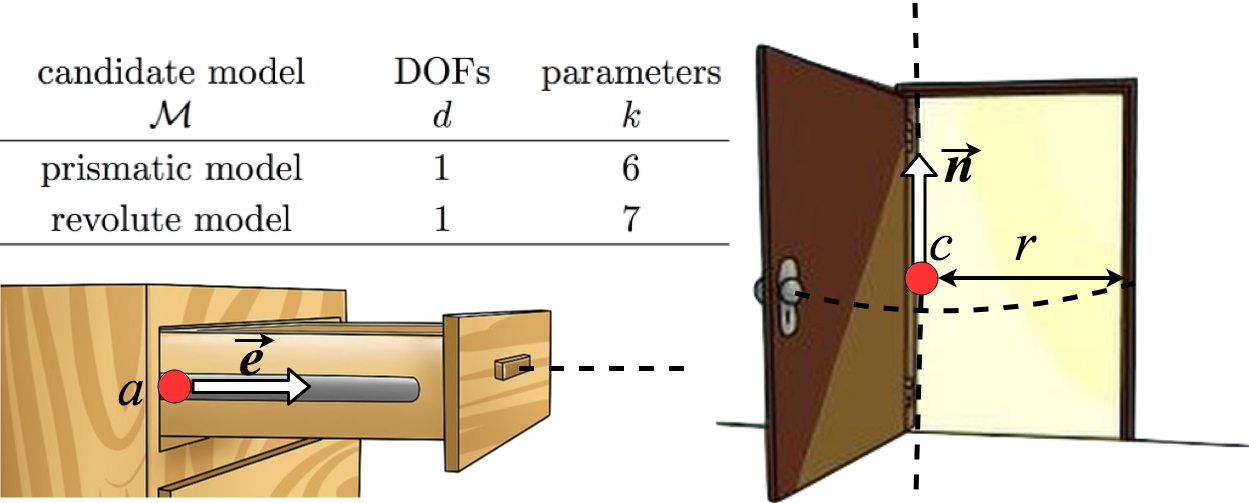}
\caption{Prismatic and revolute candidate kinematic models.}
\label{Fig7}
\end{figure}

When considering the set of doors that can be potentially operated by a service robot, their kinematic models belong to a few generic classes \citep{Sturm2012}. We have considered as candidate kinematic models a prismatic model, and a revolute model, shown in Figure \ref{Fig7}.

\subsubsection{Prismatic model}

Prismatic joints move along a single axis. Their motion describes a translation in the direction of a unitary vector $\mathbf{e}\in\mathbb{R}^{3}$ relative to a fixed origin, $\mathbf{a}\in\mathbb{R}^{3}$. The parameter vector is $\boldsymbol{\theta}=(\mathbf{a};\mathbf{e})$ with $k=6$. 

\subsubsection{Revolute model}

Revolute joints rotate around an axis that impose a one-dimensional motion along a circular arc. It can be parametrized by the center of rotation $\mathbf{c}\in\mathbb{R}^{3}$, a radius $\mathbf{r}\in\mathbb{R}$, and the normal vector $\mathbf{n}=\mathbb{R}^{3}$ to the plane where the motion arc is contained. This results in a parameter vector $\boldsymbol{\theta}=(\mathbf{c};\mathbf{n};r)$ with $k=7$. 

\vspace{-5mm}
\subsection{Model Fitting}

In the presence of noise and outliers, finding the parameter vector $\hat{\boldsymbol{\theta}}$ that maximizes the data likelihood is not trivial, as least square estimation is sensitive to outliers. The RANSAC algorithm has proven to be robust in this case and can be modified in order to maximize the likelihood. This is the approach implemented by the Maximum Likelihood Estimation SAmple Consensus (MLESAC) algorithm \citep{Torr2000}. In this case, the score is defined by the likelihood of the consensus sample. Thus, for estimating the model vector parameter $\boldsymbol{\theta}$, the log-likelihood of a mixture model is maximized \citep{Zuliani2012}:
\begin{equation}
 \hat{\boldsymbol{\theta}}=\argmax_{\boldsymbol{\theta}}{\mathcal{L}\left[e(\mathcal{D}\mid\mathcal{M},\boldsymbol{\theta})\right]}   
\end{equation}
\vspace{-1.5mm}
\begin{equation}
{\small{}
\begin{split}
\hat{\mathcal{L}}=\sum_{j=1}^{N}\log\bigg(\gamma\cdot p\left[e\mathbf{(d}_{j},\mathcal{M},\boldsymbol{\hat{\theta}})\mid\mathrm{j^{th}element\equiv\:inlier}\right] \\ + \left(1-\gamma\right)p\left[e\mathbf{(d}_{j},\mathcal{M},\boldsymbol{\hat{\theta}})\mid\mathrm{j^{th}element\equiv outlier}\right]\bigg)
\end{split}
}
\end{equation} where $\gamma$ is the mixture coefficient, which is computed with Expectation Maximization. The first and second term correspond to the error distribution $e(\mathbf{d}_j,\mathcal{M},\hat{\boldsymbol{\theta}})$ of the inliers and the outliers, respectively. The error statistics of the inliers are modeled with a Gaussian. On the other hand, the error of the outliers is described with a uniform distribution. 

\subsection{Model Selection}

Once all model candidates are fitted to the observations, the model that best explains the data has to be selected \citep{Sturm2011}. Let $\mathcal{M}_{m}$ ($m=1,...,M$) be the set of candidate models, with vector parameters $\boldsymbol{\theta}_{\mathit{m}}$. Let $p\left(\boldsymbol{\theta}_{m}\vert \mathcal{M}_{m}\right)$ be the prior distribution for the parameters. Then, the posterior probability of a given model is proportional to \citep{Hastie2009}:

\begin{equation}
{\small{}
p\left(\mathcal{M}_{m}\mid\mathcal{D}\right)\propto\int p\left(\mathcal{D}\mid\boldsymbol{\theta}_{m},\mathcal{M}_{m}\right)p\left(\boldsymbol{\theta}_{m}\mid\mathcal{M}_{m}\right)d\mathbf{\boldsymbol{\theta}}_{m}
}
\end{equation}

In general, computing this probability is difficult. Applying the Laplace approximation and assuming a uniform prior for the models, i.e. probability of model $M_m$ of being the true model before having observed any data, it can be estimated in terms of the Bayesian Information Criterion (BIC):
\begin{equation}
{\small{}
p(\mathcal{M}_{m}\mid \mathcal{D})\approx\frac{\exp\left(-\frac{1}{2}\triangle BIC_{m}\right)}{\sum_{m=1}^{M}\exp\left(-\frac{1}{2}\triangle BIC_{m}\right)}
}
\end{equation} where:{\small{} $\triangle BIC_{m}=BIC_{m}-\min\left\{BIC_{m}\right\}^{M}_1$}, and:
\begin{equation}
BIC_{m}=-2\log\left[\mathcal{L}\left(\mathcal{D}\mid\mathcal{M}_{m},\boldsymbol{\hat{\theta}}_{m}\right)\right]+k\cdot\log N
\end{equation} The first term accounts for the likelihood of the fit, and the second term for the model complexity; smaller $BIC$ are preferred. Thus, model selection can be reduced to select the model with the lowest $BIC$:
\begin{equation}
 \hat{\mathcal{M}}=\argmin_{\mathcal{M}}BIC(\mathcal{M})
\end{equation}

\subsection{Exploiting Prior Knowledge}

A robot operating in domestic environments can boost its performance by learning priors from previous experiences \citep{Calinon2016}. A small set of representative models can be used as prior information to improve the model selection and parameter estimation in an unknown environment.

Suppose that the robot has previously encountered two doors. We have two observation sequences $\mathcal{D}_{1}$ and $\mathcal{D}_{2}$, with $N_{1}$ and $N_{2}$ samples. We must choose then between two distinct models $\mathcal{M}_{1}$ and $\mathcal{M}_{2}$ or a joint model $\mathcal{M}_{1+2}$. In the first case, the posterior can be split as the two models are mutually independent:
\begin{equation}
p\left(\mathcal{M}_{1},\mathcal{M}_{2}\mid\mathcal{D}_{1},\mathcal{D}_{2}\right)=p\left(\mathcal{M}_{1}\mid\mathcal{D}_{1}\right)\cdot p\left(\mathcal{M}_{2}\mid\mathcal{D}_{2}\right)
\end{equation}

In the second case, both trajectories are explained by a single, joint model $\mathcal{M}_{1+2}$, which is estimated from the joint data $\mathcal{D}_{1}\cup\mathcal{D}_{2}$. The corresponding posterior probability is denoted $p\left(\mathcal{M}_{1+2}\mid\mathcal{D}_{1},\mathcal{D}_{2}\right)$. In order to determine whether a joint model explains the observed data better than two separate models we can compare the posterior probabilities:
\begin{equation}
   p\left(\mathcal{M}_{1+2}\mid\mathcal{D}_{1},\mathcal{D}_{2}\right)>p\left(\mathcal{M}_{1}\mid\mathcal{D}_{1}\right)\cdot p\left(\mathcal{M}_{2}\mid\mathcal{D}_{2}\right) 
\end{equation} This expression can be evaluated efficiently using the BIC
\begin{equation}
BIC\left(\mathcal{M}_{1+2}\mid\mathcal{D}_{1},\mathcal{D}_{2}\right)<\sum_{i=1}^2BIC\left(\mathcal{M}_{i}\mid\mathcal{D}_{i}\right)
\end{equation}

Intuitively, merging two models into one is beneficial if the joint model can explain the data equally well while requiring only a single set of parameters. If we consider more than two trajectories, this should be repeated for all the possible combinations. This can become hard to compute. Thus, instead, we check if merging the new data with each learned model associated with the door class being opened gives a higher posterior. In this way, when opening a refrigerator door, the observations are only going to be compared with the previous refrigerator door openings. In this way,
the observed data is more likely to match the recorded data, and trajectories
that are not likely to match (i.e. a drawer) are not considered. Finally, we pick the model with the highest posterior and record the new data, which will be used as prior knowledge for future doors. This approach is summarized in algorithm \ref{Algorithm2}.

\vspace{-1mm}
\begin{algorithm}
\small{
\caption{{\bf Model Selection Using Prior Knowledge} \label{Algorithm2}}
\small
\begin{algorithmic}
 \renewcommand{\algorithmicrequire}{\textbf{Input:}}
 \renewcommand{\algorithmicensure}{\textbf{Output:}}
\Require{New observed trajectory $\mathcal{D}_{\text{new}}=\left \{ \textbf{d}^{\text{new}}_j \right \}^{N}_1$ ;\\door class $c \in \left \{ \text{door},\; \text{cabinet door},\; \text{refrigerator door} \right \}$; \\previously observed trajectories $\mathbb{D}_c= \left \{ \mathcal{D}_s \right \} ^{S}_{1}$}

\Ensure{Best model $\mathbb{M}_{\text{best}}$and prior knowledge updated $\mathbb{D}_c$}

\State $\mathcal{M}_{\text{new}} \gets \text{Kinematic\_Model}\left(\mathcal{D}_{\text{new}}\right)$

\State $\mathbb{M}_{\text{best}}\gets\left \{ \mathcal{M}_{\text{new}} \right \}\:$, $\mathbb{D}_c\gets\mathbb{D}_c\cup\left\{\mathcal{D}_{\text{new}}\right\}$, $\: p_{\text{best}}\gets0$

\ForAll{$\mathcal{D}_s \in \mathbb{D}$}

 \State $\mathcal{M}_{s} \gets \text{Kinematic\_Model} \left (\mathcal{D}_s \right )$

 \State $\mathcal{M}_{\text{new+s}} \gets \text{Kinematic\_Model} \left ( \mathcal{D}_{\text{new}} \cup \mathcal{D}_{s} \right )$
 \small
 \If {\scriptsize{}$p\left(\mathcal{M}_{\text{new+s}}\vert\mathcal{D}_{\text{new}},\mathcal{D}_s\right)>p\left(\mathcal{M}_{\text{new}}\vert\mathcal{D}_{\text{new}}\right)p\left(\mathcal{M}_s\vert\mathcal{D}_s\right)\And p\left(\mathcal{M}_{\text{new+s}}\vert\mathcal{D}_{\text{new}},\mathcal{D}_s\right)>p_{\text{best}}$}
    \small
    \State $\mathbb{M}_{\text{best}} \gets \left\{ \mathcal{M}_{\text{new}},\;\mathcal{M}_s\right\}$
    
    \State $\mathbb{D}_{c} \gets \left\{ \mathcal{D}_{1},\ldots,\mathcal{D}_{\text{new}}\cup\mathcal{D}_s\,\ldots,\mathcal{D}_S\right\}$
      
    \State $p_{\text{best}} \gets p\left( \mathcal{M}_{\text{new+s}}\:\vert\:\mathcal{D}_{\text{new}},\mathcal{D}_s \right)$
    
    \EndIf
\EndFor
\newline
\Return $\mathbb{M}_{\text{best}}$ and $\mathbb{D}_c$
\end{algorithmic}
}
\end{algorithm}

\vspace{-4mm}
\subsection{Learning from Human Demonstrations\label{sec:Learning-for-Human}}

\begin{figure}
\centering
\includegraphics[width=0.8\linewidth]{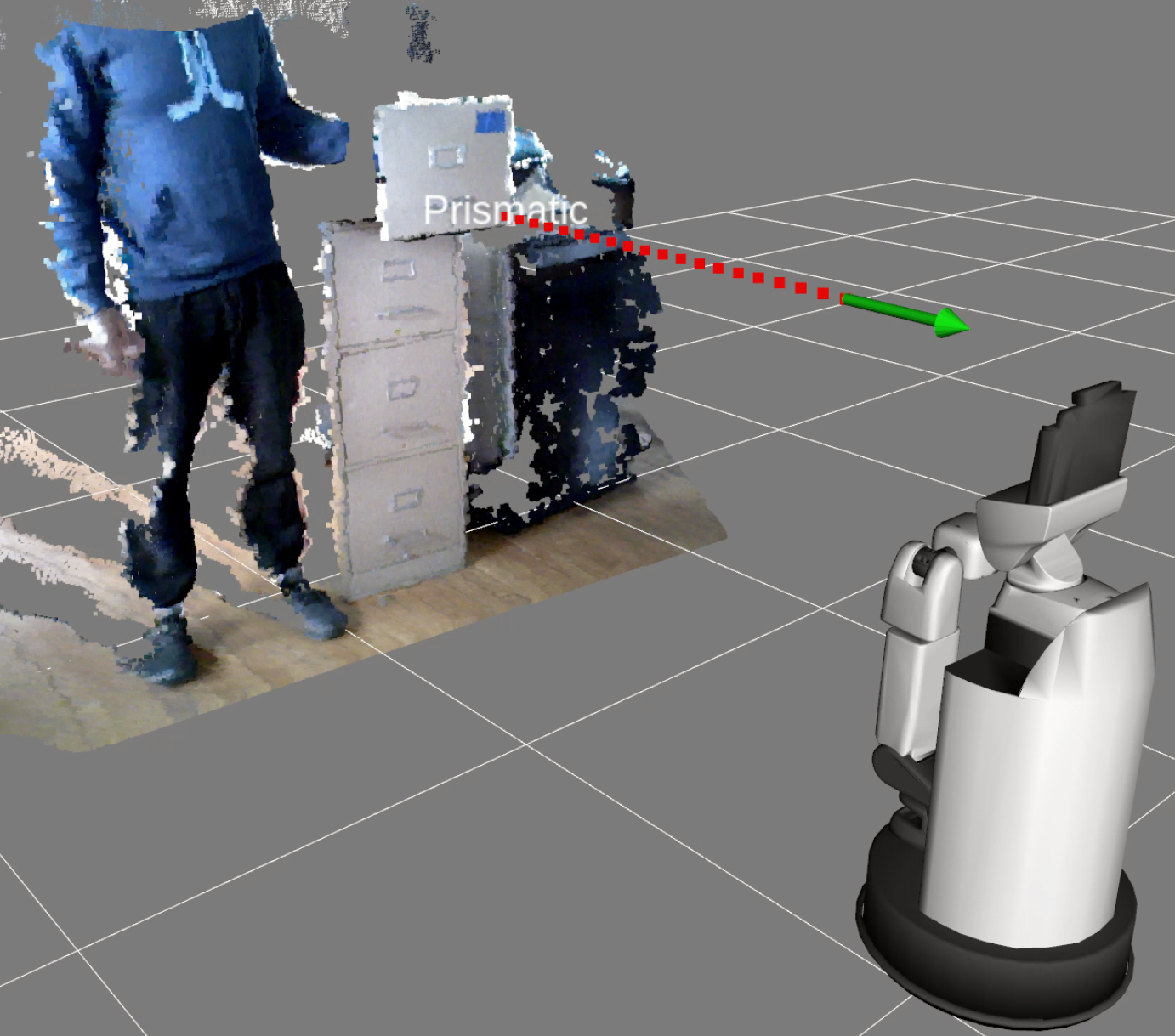}
\caption{Observations can be provided by executions of the task by a human teacher. In this case, the robot infers the motion of the cabinet is described by a prismatic model.}
\label{Fig8}
\end{figure}

If robots can learn from demonstration, this can boost the scale of the process, since nonexperts would be able to teach them \citep{Lee2017}. With our probabilistic framework, the only necessary input we need is a set of observations of the door's motion. Using our 6D-pose estimation approach, this tracking behavior can be efficiently achieved. Thus, the robot's prior knowledge can be provided by human demonstrations (Figure \ref{Fig8}).

\section{Execution of the Door Opening Motion}
\label{door_opening}

Computing the motion that enables a mobile manipulator to open a door is challenging because it requires tight coordination between arm and base. This makes the problem high-dimensional and thus hard to plan. In the previous section, we have discussed how to learn the door kinematic model from observations of its motion. In order to achieve full autonomy, although observations are not provided before-hand, the robot must also be able to operate the previously unseen door. In this section, we will discuss how these issues can be addressed through a suitable motion planning framework and an effective door opening strategy. 

\vspace{-3mm}
\subsection{Task Space Region (TSR)}

Task Space Region is a constrained manipulation planning framework presented in \citep{Berenson2011b}. The authors propose a specific constraint representation, that has been developed for planning paths for manipulators with end-effector pose constraints. The framework unifies an efficient constraint representation, constraint satisfaction strategies, and a sampling-based planner, to create a state-of-the-art whole-body manipulation planning algorithm. 

The sampling-based planner is based on rapidly exploring random trees (RRTs). Thus, it inherits many of the limitations of sampling-based methods for planning. For instance, it is very difficult to incorporate non-holonomic constraints and dynamics because these constraints would disrupt the distance metric used by the RRT. For these reasons, the applicability of this framework is limited to robots without non-holonomic constraints.

TSRs describe end-effector constraint sets as subsets of \textsl{SE}$(3)$ (Special Euclidean Group). These subsets are particularly useful for specifying manipulation tasks. Once the end-effector pose restrictions are specified in terms of a TSR, the algorithm finds a path that lies in the constraints manifold. To define a TSR, three elements are required:
\begin{itemize}
\item $\textbf{\textit{T}}^{o}_w$: Transform between the origin reference frame $o$ and the TSR frame $w$.
\item$\textbf{\textit{T}}^{w}_e$: End-effector offset transform.
\item$\textbf{\textit{B}}^{w}$: $6\times2$ matrix that defines the end-effector constraints, expressed in the TSR reference frame
\end{itemize}
\begin{equation}
    \left(\textbf{\textit{B}}^{w}\right)^T=\left(\begin{array}{cccccc}
         x_{\min} & y_{\min} & z_{\min} & \varphi_{\min} & \theta_{\min} & \psi_{\min}  \\
         x_{\max} & y_{\max} & z_{\max} & \varphi_{\max} & \theta_{\max} & \psi_{\max} 
    \end{array}\right)
\end{equation} where the first three columns bound the allowable translation along the $x$, $y$ and $z$ axes, and the last three columns bound the allowable translation assuming the Roll-Pitch-Yaw angle convention. Thus, the end-effector constraints for the considered kinematic models can be easily specified as follows.

\subsubsection{TSR representation for Prismatic Doors}

\begin{figure}
\centering
\includegraphics[width=1.0\linewidth]{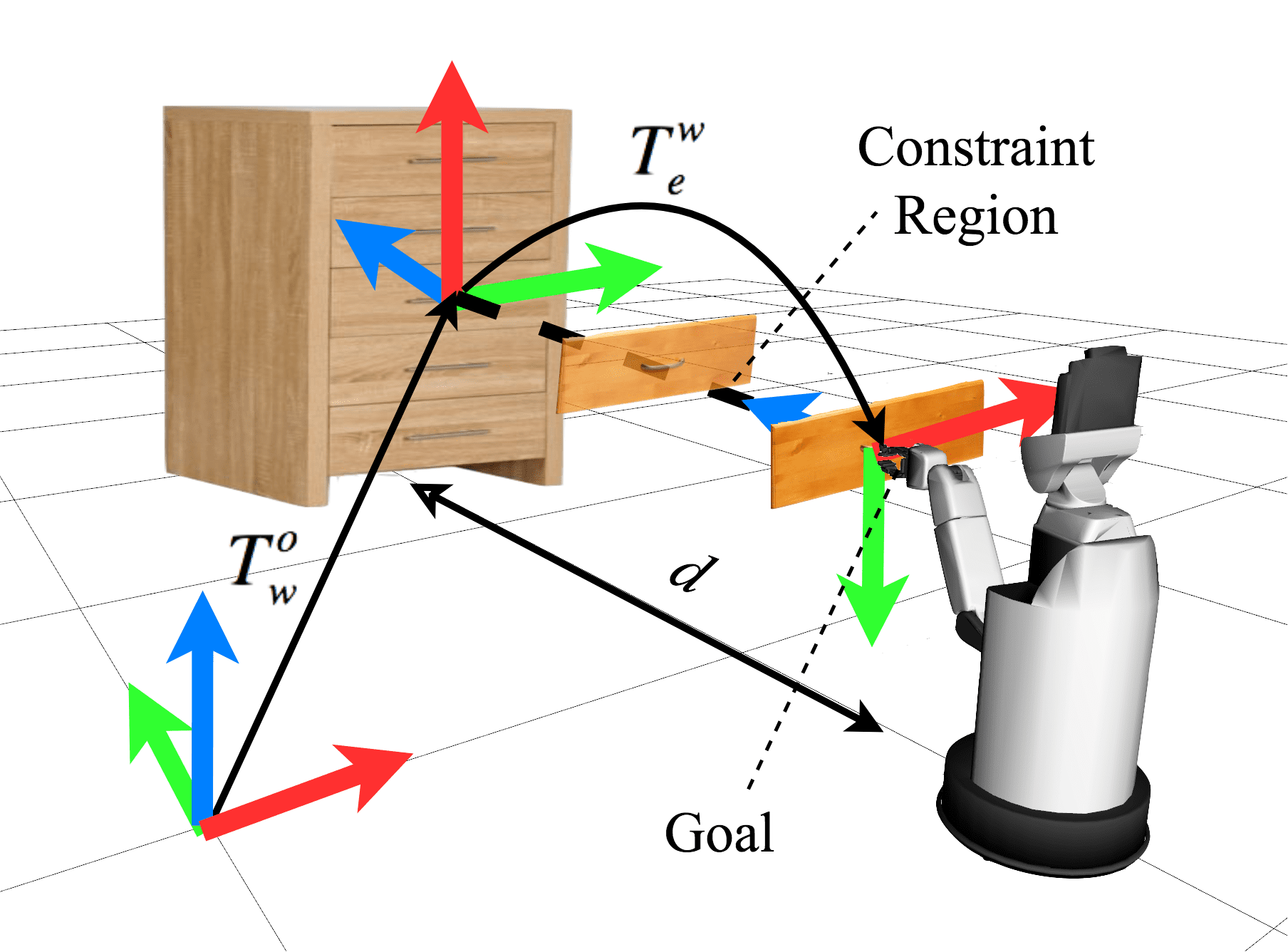}
\caption{TSR representation for operating prismatic doors. The $x$, $y$ and $z$ axis of each reference frame are red, green and blue respectively.}
\label{Fig9}
\end{figure}

By fitting a prismatic model to the observations we can estimate the axis along which the door moves. If this axis is determined, we can specify the TSR reference frame as shown in Figure \ref{Fig9}, and define the end-effector pose constraints as:

\begin{equation}
    \left(\textbf{\textit{B}}^{w}\right)^T=\left(\begin{array}{cccccc}
         \;0\;\; & 0\;\; & -d\; & 0\;\; & 0\;\; & 0\;\;  \\
         0\;\; & 0\;\; & \;0\;\; & 0\;\; & 0\;\; & 0\;\;
    \end{array}\right)
\end{equation}

\subsubsection{TSR representation for Revolute Doors}

\begin{figure}
\centering
\includegraphics[width=0.8\linewidth]{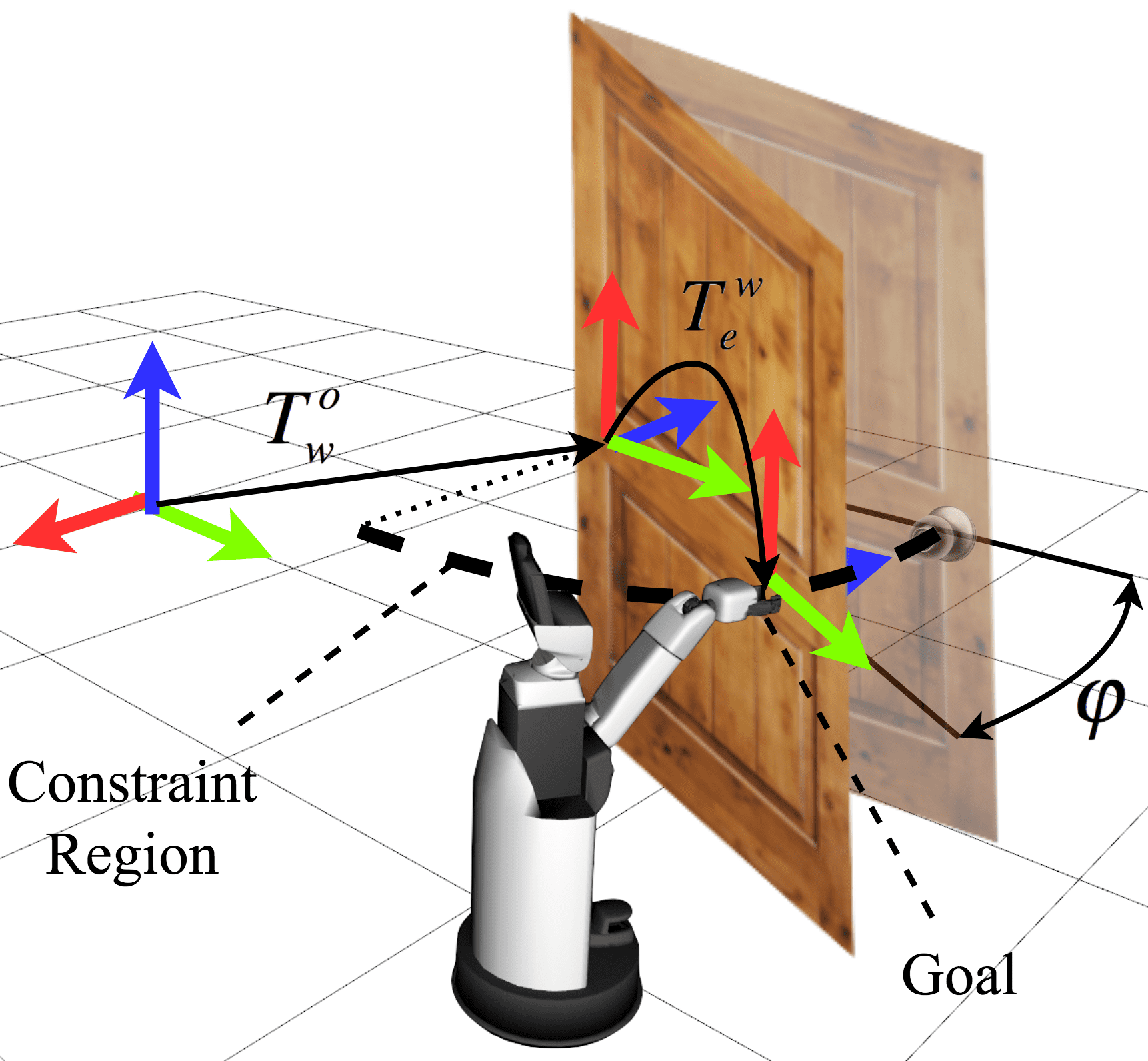}
\caption{TSR representation for operating revolute doors. The $x$, $y$ and $z$ axis of each reference frame are red, green and blue respectively.}
\label{Fig10}
\end{figure}

In the case of fitting a revolute model to the observations, we can estimate the center of rotation, the radius and the normal axis. With these parameters, we can specify the TSR reference frame as shown in Figure \ref{Fig10}, and define the end-effector pose constraints as:

\begin{equation}
    \left(\textbf{\textit{B}}^{w}\right)^T=\left(\begin{array}{cccccc}
         \;0\;\; & 0\;\; & 0\;\; & -\varphi\; & 0\;\; & 0\;\;  \\
         0\;\; & 0\;\; & 0\;\; & \;0\;\; & 0\;\; & 0\;\;
    \end{array}\right)
\end{equation}

\vspace{-6mm}

\subsection{Door Opening Procedure}

\begin{figure}
\centering
\includegraphics[width=0.96\linewidth]{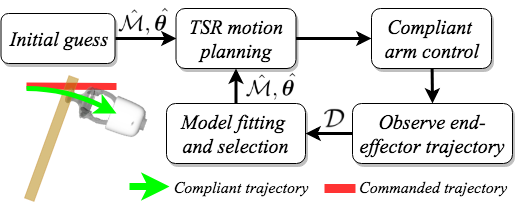}
\caption{Adaptive door opening procedure scheme. The robot opens the door following these steps iteratively.}
\label{Fig11}
\vspace{-3mm}
\end{figure}

To define the TSR reference frame, observations of the door motion are also required. Instead of using visual perception, it can also be inferred by direct actuation. Once the handle is grasped, the position of the end-effector directly corresponds with the position of the handle. As a result, the robot can make observations of the motion by solving its forward kinematics. Thus, $\mathcal{D}$ can be obtained by sampling the trajectory. We execute the door opening motion repeating iteratively a series of sequential steps, shown in Figure \ref{Fig11}. After each iteration, we re-estimate the kinematic model of the door and its parameters adding the new observations to $\mathcal{D}$. 

\begin{figure*}[h!]
\centering
\includegraphics[width=1.0\linewidth]{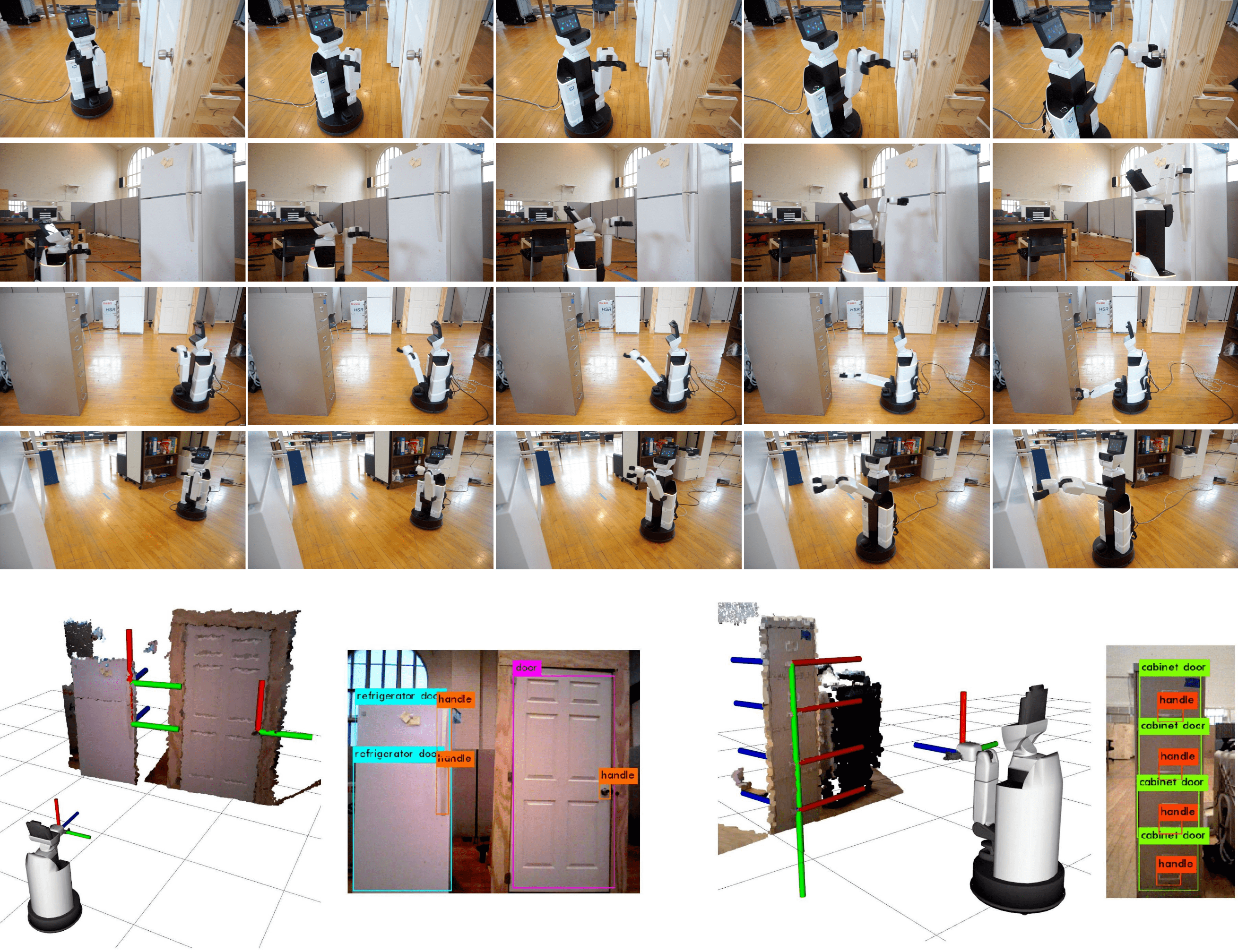}
\caption{On top, a series of pictures of the HSR robot grasping different handles, starting from various relative positions. Below, the estimated grasping pose for the handles in the scene, as well as the corresponding detections provided by our CNN. The estimated grasping pose is illustrated through the red, green and blue axis ($x$, $y$ and $z$ respectively). Note the end-effector reference frame is shown at the HSR gripper. Video demonstrations are available at https://www.youtube.com/watch?v=LbDfKPpxEss.}
\label{Fig12}
\vspace{-2mm}
\end{figure*}

\begin{figure*}[h!]
\centering
\includegraphics[width=1.0\linewidth]{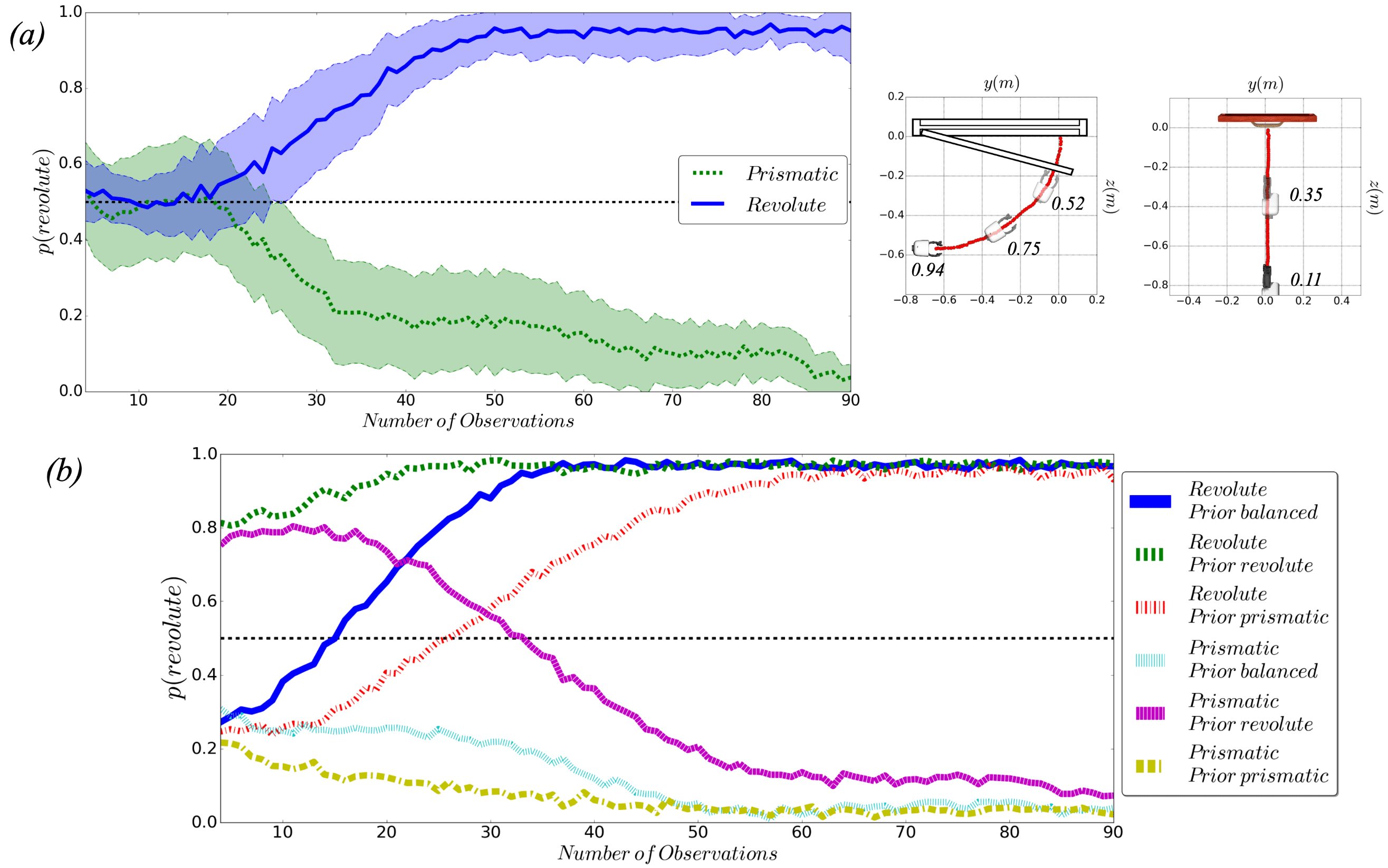}\caption{(a) The posterior of the revolute model vs the number of observations. In the legend, the door true models are indicated. The means of the executions are displayed as continuous lines. The shaded areas represent a margin of two standard deviations. Next to the plot, the evolution of the posterior along the opening trajectory is shown graphically. (b) Evolution of the revolute posterior mean against the number of observations. The legend indicates the true model of the doors being opened and the predominant prior during the realization.}
\label{Fig13}
\end{figure*}

To start the opening process, when no observations are available, we make the initial guess that the model is prismatic. Using a compliant controller, the robot's end-effector trajectory is also driven by the forces exerted by the door, adapting its motion to the true model. Thus, a certain error margin is allowed, enabling the robot to operate the door correctly even if the estimation is biased when only a few observations have been acquired.

\vspace{-4mm}

\section{Experimental Evaluation}
\label{experiments}

In order to validate experimentally the proposed door operation framework, we implemented it on the Toyota HSR robot, a robot designed to provide assistance. It is equipped with an omnidirectional wheeled base, an arm with 4 degrees-of-freedom, a lifting torso and a two-fingered gripper as an end-effector. 

In this work, we take advantage of the sensorial feedback provided by a 6-axis force sensor, located on the wrist, and an RGB-D camera, located on its head. We conducted a series of real-world experiments to test the performance of the presented grasping pose estimator and the proposed kinematic model inference process. We tested the latter in two different scenarios: with and without exploiting prior knowledge. To assess the robustness of our framework, we used different doors such as cabinet, refrigerator, and room doors with their variety of handles. Video demonstrations are available at the following \href{https://www.youtube.com/watch?v=LbDfKPpxEss}{\underline{hyperlink}}.

\vspace{-4mm}

\subsection{Grasping Pose Estimation}

For evaluating the performance of the grasping pose estimation, we focused on accuracy and speed. Regarding the door and handle detection model, due to the limited range of different doors available in the laboratory, its accuracy is best assessed, as discussed in Section \ref{detection}, by computing the mAP on the test set, with a wide variety of doors and handles. The resulting mAP of the selected model, as well as some reference values for comparison, are shown in Table \ref{tab:1}. We can see that our model's mAP is just $10\%$ lower. This performance value is close to that obtained by state-of-the-art object detectors in high-quality image datasets. 

Qualitatively, testing the model in the laboratory, the available doors and handles were effectively detected from different viewpoints. Given a successful detection, the algorithm always computed the grasping pose of the handles present in the image correctly. By solving the IK, if the handle was located within the reachable workspace, the robot was always able to grasp the handle. Using an Nvidia Geforce GTX 1080 GPU, we obtained a computation rate of $6$fps. This shows an efficient behavior of the presented real-time grasping pose estimator. In Figure \ref{Fig12} we show a series of pictures that illustrate how the HSR robot reaches the handle in different scenarios, after inferring the grasping pose with our method. An effective grasping is achieved from several starting positions and types of doors. We also show some examples of the estimated goal pose from RGB-D data. We can observe that is accurately located in the observed point cloud for all the handles simultaneously. 

\vspace{-3mm}
\renewcommand{\arraystretch}{1.5}
\begin{table}[h]
\centering
\caption{mAP comparison}
\label{tab:1}       
\begin{tabular}{||c||c||} 
 \cline{2-2}
 \multicolumn{1}{c||}{} & mAP \\
 \hline
 YOLO on COCO dataset & 55\% \\ 
 \hline
 YOLO on VOC 2012 & 58\% \\
 \hline
 YOLO on our custom dataset & 45\% \\
 \hline
\end{tabular}
\vspace{-9mm}
\end{table}

\subsection{Kinematic Model Inference}

\setcounter{figure}{15}
\begin{figure*}
\centering
\includegraphics[width=1.0\linewidth]{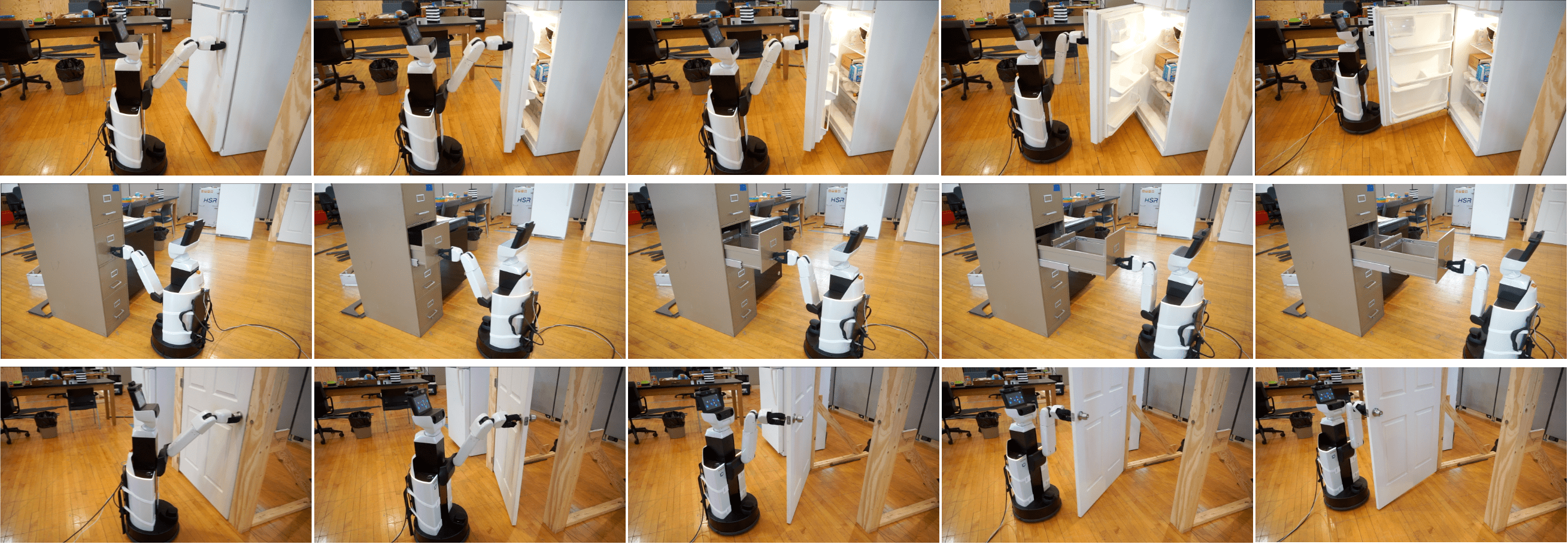}\caption{The HSR robot successfully opens different types of doors, without a priori knowledge of their kinematic model.}
\label{Fig14}
\vspace{-3mm}
\end{figure*}

\setcounter{figure}{14}
\begin{figure}[b]
\centering
\includegraphics[width=1.0\linewidth]{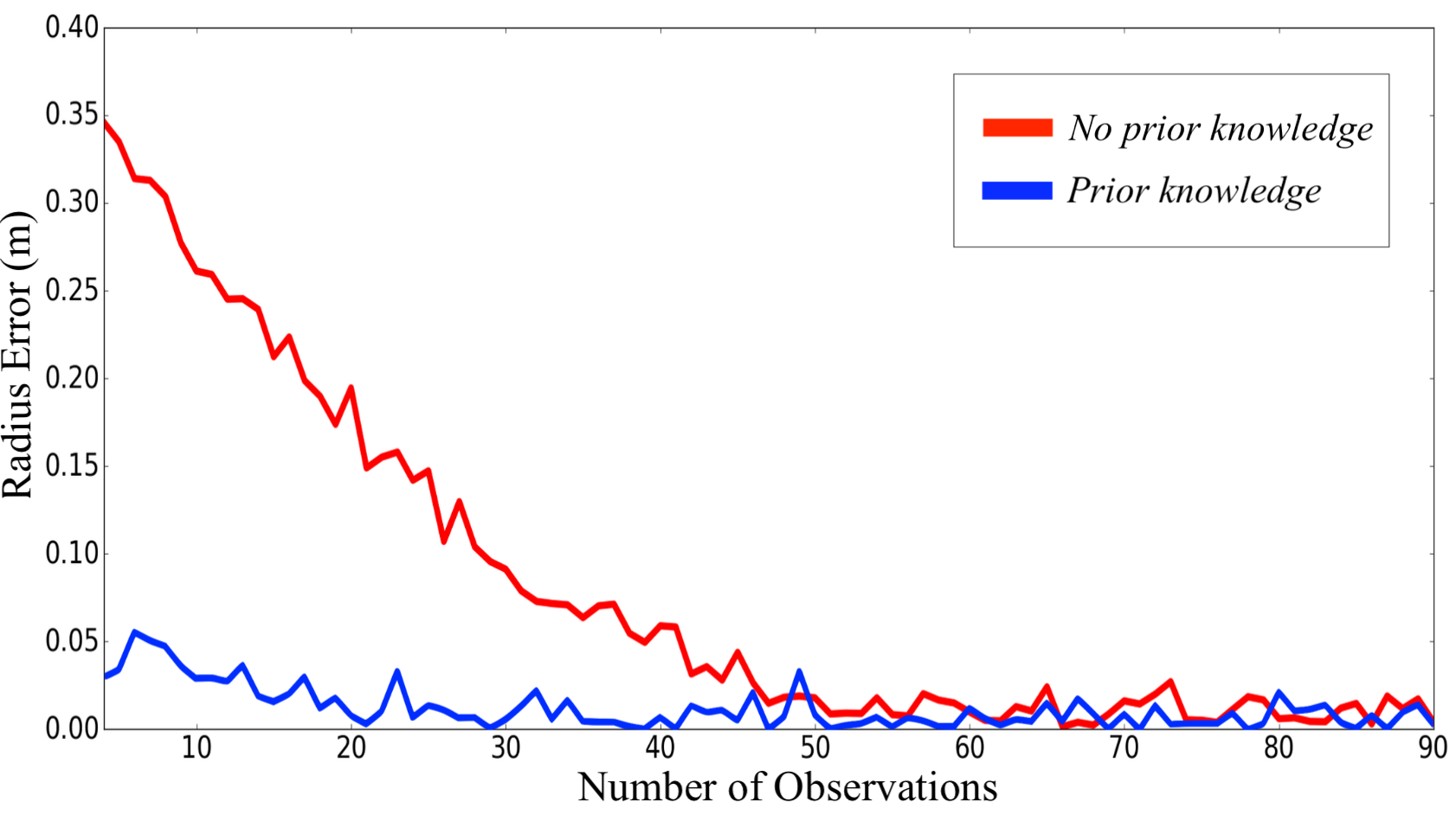}
\caption{Evolution of the estimation error of the radius of a revolute door during its operation. Two scenarios are compared: without and with previous knowledge.}
\label{Fig16}
\end{figure}

In order to evaluate the door kinematic model inference process, when no prior knowledge is available, we opened three different types of doors ten times: a drawer, a room and a refrigerator door. The task of the robot was to grasp the handle and open the door while it learned its kinematic model. The robot succeeded $26$ times out of $30$ trials ($87\%$). All four failures were due to the gripper slipping from the doorknob, most likely caused by the design of the gripper which is not very suitable to manipulate this kind of object. No errors were observed during model learning. 

We also studied the convergence of the estimators versus the number of training samples. We considered ten successful openings for each of the two considered kinematic models. Results are shown in Figure \ref{Fig13}(a). During the task, the evolution of the candidate posterior model was evaluated against the number of observations. It can be seen that the posterior probability for both cases converges towards the true model as the number of observations increases. When a  observations are acquired, the probability oscillates around $0.5$, which is consistent with considering equal priors. However, they soon diverge from this value, showing an effective behavior regarding the decision criterion. A more convergent behavior is visible in the case of a revolute door. This is due to the difference in complexity between both models. When a prismatic door is opened, the revolute model can fit the data, which does not happen in the opposite case.

\vspace{-1.5mm}

Then, we analyze our approach for exploiting prior knowledge. We reproduced the same experiments when no prior knowledge is available but for three different situations: when the prior is predominantly revolute or prismatic, and when both are balanced. Results are shown in Figure \ref{Fig13}(b). It can be observed that the behavior depends on the predominant prior. In the case it matches the true model, the posterior converges quickly. If the prior is balanced, the evolution depends on the true model. When few new observations are available, the posterior tends to converge to the simplest model which is prismatic. This is reasonable since the trajectory is very similar for both models at this point but the complexity is penalized. However, at a relatively low number of observations, the posterior rapidly converges to the true model proving, therefore, an improvement in performance. Note that priors start around $0.3$, this is because the new observations are matched to a previous model already from the starting point. Finally, in the case the prior does not match the true model, the behavior is symmetric for both doors. At the beginning, the observations converge with the predominant prior model. However, when the number of observations is sufficiently large, they converge towards the true model. A numerical evaluation of the advantage of exploiting prior knowledge is shown in Figure \ref{Fig16}, where we can observe the evolution of the radius estimation error when opening a revolute door with and without providing demonstrations of its motion. By exploiting prior knowledge, we can see that the estimation error is almost null from the initial stages of the opening process, which does not occur in the other scenario. Finally, in Figure \ref{Fig14} we show a series of pictures that illustrate how the HSR successfully opens different doors. Combining the proposed probabilistic approach, with the TSR manipulation framework, the robot can operate doors autonomously in an unknown environment.

\section{Conclusion}
\label{conclusions}

In this work, our objective is to push the state-of-the-art towards achieving autonomous door operation. The door opening task involves a series of challenges that have to be addressed. In this regard, we have discussed the detection of doors and handles, the handle grasp, the handle unlatch, the identification of the door kinematic model, and the planning of the constrained opening motion.

The problem of rapidly grasping door handles leads to the first paper contribution. A novel algorithm to estimate the required end-effector grasping pose for multiple handles simultaneously, in real-time, based on RGB-D has been proposed. We have used a CNN, providing reliable results, and efficient point cloud processing to devise a high-performance algorithm, which proved robust and fast in the conducted experiments. Then, in order to operate the door reliably and independently of its kinematic model, we have devised a probabilistic framework for inferring door models from observations at run time, as well as for learning from robot experiences and from human demonstrations. By combining the grasp and model estimation processes with a TSR robot motion planner, we achieved a reliable operation for various types of doors. 

Our desire is to extend this work to include more general and complex kinematic models \citep{Barragan2014,Hoefer2014}. This would enable robots, not only to achieve robust door operations but would ultimately achieve generally articulated object manipulation. Furthermore, the use of non-parametric models, such as Gaussian processes, would allow the representation of even more complex mechanisms. Also, we would like to explore in more depth the possibility of integrating our system in a general Learning from Demonstration (LfD) framework.


%
%

\bibliographystyle{abbrvnat}
\bibliography{Referencias}   

\end{document}